\documentclass[11pt]{article}
\usepackage[margin=1in]{geometry}
\usepackage{graphicx,booktabs,amsmath,amssymb,xcolor,hyperref,natbib,pifont}
\usepackage[T1]{fontenc}
\hypersetup{colorlinks=true,linkcolor=blue,citecolor=blue,urlcolor=blue}
\title{\textbf{How Modular Is a Frontier Mixture-of-Experts?\\
A Pre-registered Causal Test in Which Apparent Expert Modularity Mostly Dissolves}}
\author{
  Tony Salomone \\ Transformer Lab \\ \texttt{tony.salomone@lab.cloud}
  \and
  Deep Gandhi \\ Transformer Lab
  \and
  Ali Asaria \\ Transformer Lab
}
\date{\today}

\begin{document}
\maketitle

\begin{abstract}
Sparse Mixture-of-Experts (MoE) models route each token to a few of many experts, inviting the
hypothesis that experts form \emph{functional modules} tied to capabilities or languages. We test
this causally on \textbf{Command~A+}, a frontier open-weights MoE (218B total / 25B active; 128
experts, 8 active, +1 shared)~\citep{commandaplus2026}. We build a routing-mass \emph{atlas},
\textbf{pre-register} six family$\to$axis hypotheses before any intervention, and ablate each family
at inference time against a size-matched random-expert null, measuring whether it \emph{selectively}
breaks its own axis (worst off-target effect $\leq\tfrac13$ of on-target). Crucially, we test the same families under four metrics and a held-out,
independent-corpus run with bootstrap confidence intervals. Our finding is cautionary: \textbf{robust
functional modularity is rare and measurement-dependent}. Of six pre-registered families, only
\emph{one}, the Arabic-language family, is a clean selective module that survives an independent
corpus and a conservative statistical bar (1/6; a more permissive pre-registered point rule admits
3/6, but that count is threshold-sensitive: es clears selectivity by only $0.002$ and code misses by
$0.009$, so families straddle the boundary from both sides). Every other family has a real causal effect
yet fails selectivity, and its apparent modularity \emph{flips with the measurement}: with the
\textbf{corpus} (Spanish is selective on one corpus but bleeds into Arabic on a second), with the
\textbf{metric} (math is entangled with general reasoning under task accuracy yet looks selective
under solution-likelihood), and with the \textbf{statistical bar} (the very 1/6-vs-3/6 count). A positive control on Qwen3-30B-A3B recovers its published disjoint structure,
confirming the method detects modularity when present (a sensitivity check, with no negative control).
The verdict reproduces on the
un-quantized BF16 model, ruling out a 4-bit quantization artifact. We conclude that ablation-based
modularity verdicts are \emph{not safe} unless the corpus, metric, and statistical bar are
controlled, and that, in Command~A+ so controlled, only one of six pre-registered families is a
robust module. We release the atlas and
ablation data.
\end{abstract}

\section{Introduction}
A sparse MoE activates a handful of its experts per token (here, 8 of 128), so most expert weight is
cold at any moment. This sparsity makes a tempting structural story: perhaps experts specialize into
\emph{families} that each own a capability (math, code) or a language, so that a frontier MoE is
\emph{functionally modular}. Modularity, if real, would be a powerful lens for interpretability,
safety, and editing.

Modularity is a \emph{causal} claim, and the field has repeatedly found that observational
signals, which experts a token is routed to, do not by themselves establish what the output depends
on~\citep{causalaudit2026,counterfactual2024}. We therefore ask the question the way it must be
answered: \emph{does ablating a family of experts selectively break its hypothesized axis while
sparing the others?} We make three commitments: (1) \textbf{pre-registration}, families are defined
from routing and their hypothesized roles frozen before any ablation; (2) a \textbf{size-matched
random null}, an effect counts only if it beats ablating the same number of random experts; and
(3) a \textbf{falsifiable selectivity criterion}, with ``not modular'' an admissible outcome.

Our central methodological move is to treat the \emph{measurement} as a variable. We score the same
ablations under four metrics (task accuracy, problem-text likelihood, solution likelihood, and
per-language likelihood on a held-out independent corpus) and report \emph{bootstrap confidence
intervals} under both a permissive and a conservative decision rule. The result is sobering and, we
argue, the contribution: \textbf{of six pre-registered families, only the Arabic family is a robust
module}; every other family sits at the decision boundary and its apparent modularity is created or
destroyed by the choice of corpus, metric, or statistical bar. We localize one genuine module, and
we show that in this model the other apparent modules dissolve once the measurement is controlled.

\textbf{Contributions.} (1) A pre-registered, control-anchored causal protocol for expert modularity
with explicit corpus/metric/statistical robustness. (2) One clean positive result: a causally
localized Arabic-language module. (3) A \emph{threshold-independent} result that routing mass predicts
neither causal importance nor selectivity: the high-routing-mass shared-core is causally redundant,
and high-lift families are entangled rather than modular. (4) A cautionary demonstration, with
mechanism for each failure mode, that in this model apparent modularity is fragile to the choice of
corpus, metric, and statistical bar, and mostly dissolves under rigorous testing.

\section{Related work}
Routing analyses map expert specialization observationally~\citep{multiroute2025,
translspecialists2026, expertstrikesback2026}; \citet{multiroute2025} report that Qwen3-class MoEs
form \emph{disjoint} language vs.\ task expert sets, which we use as a positive control.
\citet{expertstrikesback2026} argue experts are fine-grained task specialists rather than broad
modules. On the causal side, \citet{causalaudit2026} show no observational routing metric predicts
ablation importance; \citet{knowattr2026} find MoE knowledge distributed and robust to component
removal; \citet{counterfactual2024} warn that under redundancy ablation has low recall.
\citet{cmasurvey2024} and \citet{circuithyptest2024} formalize interventional tests and
matched-random nulls, which we adopt. \citet{micro2026} show clean modularity when it is
\emph{trained in}; we test whether it \emph{emerges} in a frontier model. Our delta is a
pre-registered causal test on a 218B MoE that explicitly varies corpus, metric, and statistical bar,
which prior work does not.

\section{Method}
\paragraph{Model and ablation operator.} Command~A+ is served on a single RunPod B200 via an
in-process vLLM engine (the full study used under 30 B200-hours). We study the open-weights W4A4
(NVFP4) build (a quantization caveat we return to in \S6). We ablate a set of experts by \textbf{router masking with gate renormalization}:
at every MoE layer we set the masked experts' router logits to $-\infty$ before the top-$k$ softmax.
A smoke test confirms the mechanism is exact (masked experts go from thousands of live routings to
\emph{zero}; leak $=0$).

\paragraph{Atlas and pre-registration.} We drive a labeled probe corpus through the router, record
per-(layer,\,expert) routing mass, and pre-register each family as its top-16 experts by lift,
excluding the always-on shared experts. Six families are frozen before any ablation: math, code,
general; Arabic (ar), Chinese (zh), Spanish (es).

\paragraph{Causal test and decision rules.} For family $F$ with axis $A$, we ablate $F$ and score all
axes. A family is \emph{modular} iff (i) its on-target effect beats the size-16 random-expert null
(mean${}+2\sigma$) \emph{and} (ii) its worst off-target effect is $\leq\tfrac13$ of on-target
(selectivity). We report \textbf{two rules}: a \emph{point} rule on the raw effect sizes (the
pre-registered form), and a \emph{conservative} rule using bootstrap 95\% CIs (beats-null requires
the lower CI of on-target to exceed the null; selectivity requires the upper CI of the worst
off-target to fall below one third of the lower CI of on-target). The null itself is estimated from
$\sim$10 random draws, so mean${}+2\sigma$ is a noise band, not a calibrated $p$-value. The
$\tfrac13$ selectivity ratio and the null construction were pre-specified (fixed from separate
control-band runs before the ablation deltas were observed).

\paragraph{Four metrics, two language corpora.} Because the verdict turns out to depend on \emph{how}
breakage is measured, we score the same ablations under: task \textbf{accuracy} (MATH-500 $n{=}25$,
HumanEval $n{=}60$, MMLU-Pro $n{=}80$, single seed; small slices, so individual accuracy deltas carry
item-sampling noise, binomial SE $\approx 0.03$--$0.10$); \textbf{problem-text NLL}; \textbf{solution NLL} (teacher-forced likelihood of the gold
answer); and \textbf{per-language NLL} on two \emph{distinct} corpora, the \textbf{FLoRes-200}
passages (accessed via Belebele, which is built on them) and an independent \textbf{Wikipedia} corpus
of natural articles. The Wikipedia run, with bootstrap CIs, is the one we rely on for the headline verdict; comparing it against
the FLoRes-200 passages is itself a cross-corpus robustness test.

\paragraph{Positive control.} We run the identical pipeline on Qwen3-30B-A3B (same 128/8 shape),
where the literature reports clean language/task families, to verify the detector is sensitive. This
single control establishes \emph{sensitivity} (the method finds modularity that is present), not
\emph{specificity}; a negative control is left to future work.

\section{Results}

\begin{figure}[t]\centering
\includegraphics[width=0.6\linewidth]{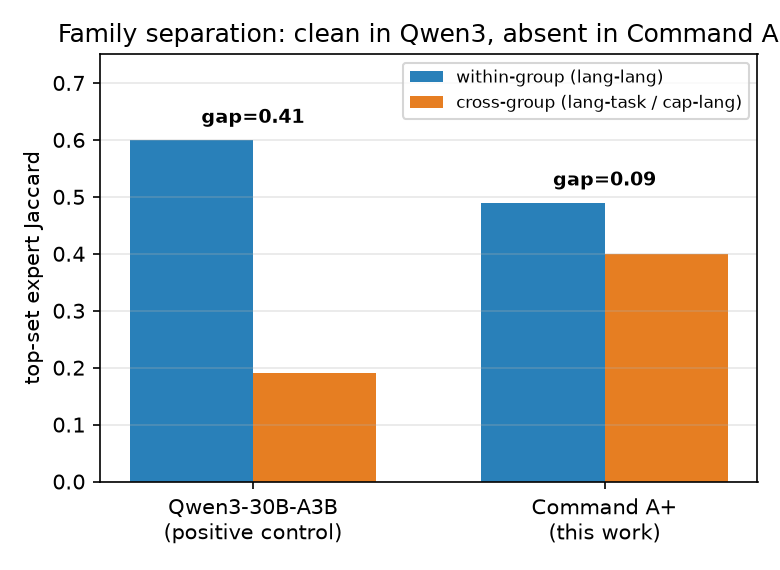}
\caption{Family separation. Qwen3 (positive control) shows a large within-vs-cross-group gap (clean
families); Command~A+ shows almost none. This is a \emph{directional} comparison, not a matched
effect size: the contrasts (language--task vs.\ capability--language) and layer counts (48 vs.\ 32)
differ (see \S4).}
\label{fig:sep}
\end{figure}

\paragraph{The detector works; Command~A+ routing is weakly separated (Fig.~\ref{fig:sep}).} On
Qwen3-30B-A3B our pipeline recovers the published structure: language--language top-set Jaccard
$0.60$ vs.\ language--task $0.19$ (gap $0.41$). The same analysis on Command~A+ yields a separation
gap of only $\mathbf{0.09}$ (language--language $0.49$, capability--language $0.40$). This is a
\emph{directional} comparison, not a matched effect size: the Qwen3 contrast is language vs.\
\emph{task} while ours is capability vs.\ language, on models of different depth ($48$ vs.\ $32$
layers) and different probe corpora. Routing does not factor into clean families, and coverage is
only mildly concentrated ($\sim$64 of 128 experts carry $\sim$62\% of mass, vs.\ $50\%$ under uniform routing). The detector finds structure when present; Command~A+ has little.

\begin{figure}[t]\centering
\includegraphics[width=0.6\linewidth]{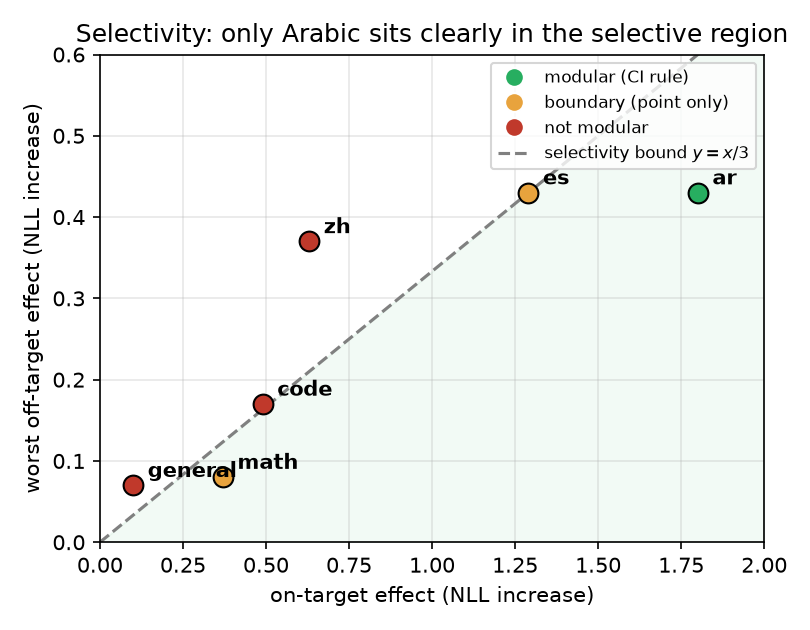}
\caption{Selectivity scatter. Each family's on-target effect (x) vs.\ its worst off-target effect
(y); points below the dashed $y=x/3$ line are selective. Only Arabic sits clearly in the selective
region; es, math, and code cluster on the boundary.}
\label{fig:scatter}
\end{figure}

\paragraph{The hardened verdict: one survivor (Table~\ref{tab:verdict}, Fig.~\ref{fig:scatter}).}
On the held-out independent corpus with bootstrap CIs, the conservative rule confirms \textbf{exactly
one} modular family, Arabic; the permissive point rule admits three (ar, es, math), but the boundary
is crowded: es clears selectivity by just $0.002$ and the \emph{rejected} code family misses by
$0.009$, so two families straddle it from opposite sides and the 3/6 count is threshold-sensitive. Of
the three admitted, only Arabic also survives the conservative CI rule: es fails it (and proves
corpus-dependent, below), while math, though it clears the point rule by a comfortable $0.04$,
collapses under the bootstrap CIs and reverses under task accuracy (below), the very fragility the
conservative rule exists to catch. Every family except \emph{general} has a
real causal effect: the question is always \emph{selectivity}, not whether the experts matter.

\begin{table}[t]\centering
\caption{Hardened verdict (independent corpus, per-item NLL, bootstrap 95\% CIs). On-target effect is
the NLL increase on the family's own axis; ``worst off'' is the largest within-domain off-target
increase. Point rule = raw effect sizes (pre-registered); CI rule = conservative bootstrap bounds.
These NLL legs use $n{=}80$--$150$; the task-accuracy results discussed in the text (e.g.\ the
math$\leftrightarrow$general entanglement) use smaller slices (MATH-500 $n{=}25$, HumanEval $n{=}60$,
MMLU-Pro $n{=}80$, single seed), without CIs.}
\label{tab:verdict}
\small
\begin{tabular}{llccccc}
\toprule
Family & Domain & On-target [95\% CI] & Null $\mu{+}2\sigma$ & Worst off & Point rule & \textbf{CI rule} \\
\midrule
\textbf{ar} & lang & \textbf{1.80} [1.73, 1.86] & 0.64 & 0.43 (en) & \checkmark & \textbf{\checkmark\ modular} \\
es      & lang & 1.29 [1.23, 1.36] & 0.58 & 0.43 (ar) & \checkmark\ {\footnotesize(by 0.002)} & \ding{55} \\
zh      & lang & 0.63 [0.59, 0.67] & 0.42 & 0.37 (en) & \ding{55} & \ding{55} \\
math    & cap  & 0.37 [0.33, 0.41] & 0.15 & 0.08 (gen) & \checkmark & \ding{55} \\
code    & cap  & 0.49 [0.39, 0.61] & 0.08 & 0.17 (math) & \ding{55}\ {\footnotesize(by 0.009)} & \ding{55} \\
general & cap  & 0.10 [0.05, 0.16] & 0.19 & --- & \ding{55} & \ding{55} \\
\bottomrule
\end{tabular}
\end{table}

\paragraph{The one robust module: Arabic.} Ablating the Arabic family raises Arabic per-token NLL by
$+1.80$ nats [95\% CI $1.73$, $1.86$], a $\sim$6$\times$ rise in perplexity (from $8.2$ to $49$),
against a random-null band of only $0.64$ nats and on a corpus it was never tuned to, while its worst
off-target (English, $+0.43$) stays well under the selectivity bar. By every standard we applied,
Arabic is a clean, causally localized language module.

\begin{figure}[t]\centering
\includegraphics[width=0.72\linewidth]{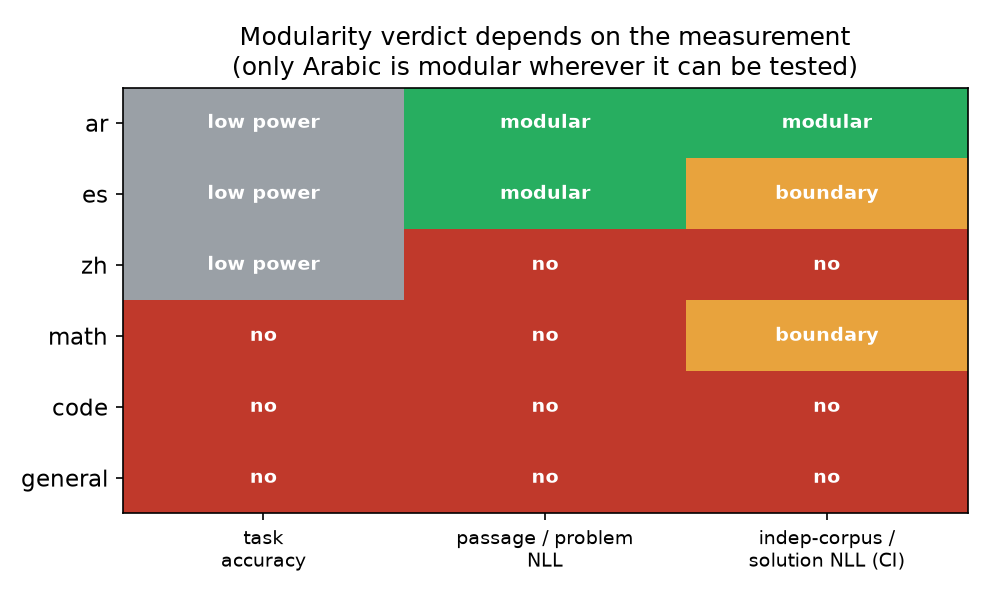}
\caption{The verdict depends on the measurement. Each cell is a family's modularity verdict under a
given metric/corpus (green = modular, amber = boundary, red = no; grey = the metric had no power to
test that cell, i.e.\ the near-ceiling Belebele accuracy for languages). Only Arabic is modular
wherever it can be tested.}
\label{fig:flip}
\end{figure}

\paragraph{Why the rest fail: measurement-dependence with mechanism (Fig.~\ref{fig:flip}).} Each
non-survivor fails along a different axis of fragility. \textbf{Spanish is corpus-dependent}: clean on
the FLoRes-200 passages, but on independent Wikipedia text ablating Spanish bleeds $+0.43$ into
Arabic, right at threshold. The es and ar families overlap in only 2 of their 16 experts
(\{30, 86\}; Jaccard $0.067$); we report the bleed but do not isolate whether those shared experts or
broader functional entanglement drive it. \textbf{Math is metric-dependent}: entangled with general
reasoning under task accuracy (ablating math drops MATH by $0.16$ \emph{and} MMLU-Pro by $0.11$, on
the small accuracy slices, $n{=}25$/$80$), apparently selective under solution-likelihood point
estimates, and non-selective once CIs are applied. The math$\leftrightarrow$general entanglement is a
\emph{shared-reasoning} effect: it appears only when the off-target metric engages reasoning (task
accuracy) and recedes when it does not (likelihood of a short answer). \textbf{Code is
metric-dependent the other way}: no detectable accuracy effect ($0.00$ on HumanEval) yet the largest
problem-text NLL effect of any capability ($+1.29$), heavily entangled (bleeding $+0.62$ into math).
The accuracy null is itself $n{=}60$, single seed, underpowered to separate a small real effect from
true redundancy. So we can say the code experts demonstrably \emph{process} code text and are
\emph{not selective} (the NLL bleed is large); whether they are dispensable for \emph{solving} is not
established by the underpowered $0.00$ accuracy null.

\paragraph{The result is not a quantization artifact (BF16).} To rule out that 4-bit quantization
manufactures the apparent non-selectivity (a false negative), we re-ran the identical consolidation
ablations on the \textbf{un-quantized BF16 base model} (8$\times$A100, tensor-parallel). The verdict
reproduces: under the conservative CI rule, BF16 again confirms \textbf{exactly one} modular family,
Arabic (on-target $+1.81$ nats vs.\ W4A4's $+1.80$; null band $0.69$), while zh, math, code, and
general remain non-selective with effect sizes within a few hundredths of the W4A4 values. The only
family to move is the knife-edge es, whose point-rule selectivity flips from just passing in W4A4
($+0.002$) to just failing in BF16 ($-0.015$), consistent with its sitting exactly on the decision
boundary. The negative finding therefore holds at full precision: it is a property of the model, not
of quantization.

\paragraph{Routing mass does not predict modularity.} The routing-defined ``universal'' shared-core
is causally redundant rather than load-bearing: a dose-response masking up to 32 of the most-shared
experts degrades MMLU-Pro by only $\sim$0.11, gracefully and with no cliff, the signature of
distributed, redundant computation.
Family lift correlates loosely with causal \emph{importance} (the two highest-lift families, math and
Arabic, have the strongest on-target effects) but not with \emph{selectivity}: high-lift code and
math families are entangled.

\section{Discussion}
A pattern runs through the results: \emph{representational} axes separate more cleanly than
\emph{computational} ones. Arabic, the one robust module, is a surface property of the token
distribution (a distinct script and lexicon) that routing can factor out. The capability axes
(math, code, general) are computations that share machinery, so they entangle, and they entangle
\emph{visibly only under metrics that exercise the shared computation}. But even representational
modularity is fragile: Spanish, also a language, fails on a held-out corpus because it shares experts
with Arabic. The practical lesson for interpretability is concrete: an ablation-based modularity
claim is not safe unless it is shown to survive an independent corpus, more than one metric, and a
conservative statistical bar. So tested, only one of the six families in this model qualifies as a
clean module. We make no base-rate claim about MoEs in general from a single model; what generalizes
is the methodological requirement, not the count.

\section{Limitations and future work}
\textbf{Quantization (addressed).} A natural worry is that 4-bit quantization blurs expert
specialization and thereby manufactures the apparent non-selectivity we report, as a false negative.
We tested this directly with the BF16 re-run above: the verdict reproduces at full precision (CI rule
1/6, Arabic; effects within hundredths of the W4A4 values), so the negative finding is a property of
the model, not of quantization. Two structural points corroborate this: only expert FFN weights are
quantized (the \textbf{router}, and hence the family boundaries we ablate, is full precision), and
selectivity appears condition-dependently rather than being uniformly smeared. We still note the
result is established on these two precisions, not exhaustively across quantization schemes. \textbf{Power and corpus.} The NLL legs use
$n\!=\!80$--$150$ per axis, but the task-accuracy legs (which the math$\leftrightarrow$general
entanglement rests on) are smaller: MATH-500 $n{=}25$, HumanEval $n{=}60$, MMLU-Pro $n{=}80$. All are
single seed (bootstrap CIs mitigate but do not replace multiple seeds), and the random null is
estimated from $\sim$10 draws. The language leg uses two corpora (FLoRes-200 passages and Wikipedia),
but a wider corpus sweep would further test corpus-dependence. \textbf{Coverage.} The
retrieval/agentic and safety capability families are untested, and with one surviving module we
cannot say \emph{which} languages modularize or why Arabic is special. \textbf{Single model.} Our
empirical conclusions are about Command~A+ (plus one positive-control model); whether they hold
across MoE families is open. What we offer as general is the \emph{methodological} requirement that
modularity claims be measurement-controlled, not a base rate for how often clean modules occur.
\textbf{Atlas.} Families are
routing-defined; a \textbf{causal-attribution atlas}, \textbf{injection} probes, and
\textbf{large-fraction} ablation are the natural ways to probe what 16/128 routing ablation cannot.

\section{Conclusion}
Robust functional modularity is rare in the frontier MoE we tested. Of six pre-registered expert
families, only the Arabic language family is a clean, selective module that survives an independent
corpus and a conservative bar; every other family sits at the decision boundary and its apparent
modularity is made or unmade by the corpus, the metric, or the statistical treatment. We localize one
genuine module and show that the rest dissolve under control. Our empirical verdict is about
Command~A+; what we offer for MoEs in general is the methodological lesson, that modularity claims
must be measurement-controlled to be trusted. We
release our atlas and ablation data and point to causal-attribution atlases and injection probes as
the way to push the question further.

\section*{Availability}
We release the expert atlas (routing-mass matrix and the frozen pre-registered family map) and the
ablation result data, together with an evidence ledger tracing every reported number to its source,
at \url{https://github.com/transformerlab/exp-command-a-plus-moe-modularity} under CC~BY~4.0. The
router-logging, masking, and evaluation harness is available from the authors on request. We do not
redistribute the model: Command~A+ is openly available under Apache-2.0~\citep{commandaplus2026}, and
we pin the exact revision used.

\bibliographystyle{plainnat}
\bibliography{references}

\begin{thebibliography}{10}
\providecommand{\natexlab}[1]{#1}
\providecommand{\url}[1]{\texttt{#1}}
\expandafter\ifx\csname urlstyle\endcsname\relax
  \providecommand{\doi}[1]{doi: #1}\else
  \providecommand{\doi}{doi: \begingroup \urlstyle{rm}\Url}\fi

\bibitem[AlKhamissi et~al.(2026)AlKhamissi, De~Sabbata, Tuckute, Chen,
  Schrimpf, and Bosselut]{micro2026}
Badr AlKhamissi, C.~Nicol\`o De~Sabbata, Greta Tuckute, Zeming Chen, Martin
  Schrimpf, and Antoine Bosselut.
\newblock Mixture of cognitive reasoners: Modular reasoning with brain-like
  specialization.
\newblock \emph{arXiv preprint arXiv:2506.13331}, 2026.

\bibitem[Bandarkar et~al.(2025)Bandarkar, Yang, Fayyaz, Hu, and
  Peng]{multiroute2025}
Lucas Bandarkar, Chenyuan Yang, Mohsen Fayyaz, Junlin Hu, and Nanyun Peng.
\newblock Multilingual routing in mixture-of-experts.
\newblock \emph{arXiv preprint arXiv:2510.04694}, 2025.

\bibitem[{Cohere}(2026)]{commandaplus2026}
{Cohere}.
\newblock Command a+.
\newblock \url{https://cohere.com/blog/command-a-plus}, 2026.
\newblock Open-weights sparse Mixture-of-Experts model (218B total / 25B
  active; 128 experts, 8 active, +1 shared), Apache-2.0.
  \texttt{CohereLabs/command-a-plus-05-2026}.

\bibitem[Engmann et~al.(2026)Engmann, Adriano, and Giese]{causalaudit2026}
Leonard Engmann, Christian~Medeiros Adriano, and Holger Giese.
\newblock From observation to intervention: A causal audit of expert importance
  in mixture-of-experts models.
\newblock \emph{arXiv preprint arXiv:2606.10703}, 2026.

\bibitem[Herbst et~al.(2026)Herbst, Wermter, and Lee]{expertstrikesback2026}
Jeremy Herbst, Stefan Wermter, and Jae~Hee Lee.
\newblock The expert strikes back: Interpreting mixture-of-experts language
  models at expert level.
\newblock \emph{arXiv preprint arXiv:2604.02178}, 2026.

\bibitem[Martin et~al.(2026)Martin, Bandarkar, and Peng]{translspecialists2026}
Liu~O. Martin, Lucas Bandarkar, and Nanyun Peng.
\newblock Extracting small translation specialists from llms by aggressively
  pruning experts.
\newblock \emph{arXiv preprint arXiv:2605.28042}, 2026.

\bibitem[Mueller et~al.(2024{\natexlab{a}})Mueller, Brinkmann, Li, Marks, Pal,
  Prakash, Rager, Sankaranarayanan, Sen~Sharma, Sun, Todd, Bau, and
  Belinkov]{cmasurvey2024}
Aaron Mueller, Jannik Brinkmann, Millicent Li, Samuel Marks, Koyena Pal, Nikhil
  Prakash, Can Rager, Aruna Sankaranarayanan, Arnab Sen~Sharma, Jiuding Sun,
  Eric Todd, David Bau, and Yonatan Belinkov.
\newblock The quest for the right mediator: Surveying mechanistic
  interpretability through the lens of causal mediation analysis.
\newblock \emph{arXiv preprint arXiv:2408.01416}, 2024{\natexlab{a}}.

\bibitem[Mueller et~al.(2024{\natexlab{b}})]{counterfactual2024}
Aaron Mueller et~al.
\newblock Missed causes and ambiguous effects: Counterfactuals pose challenges
  for interpreting neural networks.
\newblock \emph{arXiv preprint arXiv:2407.04690}, 2024{\natexlab{b}}.

\bibitem[Shi et~al.(2024)Shi, Beltran-Velez, Nazaret, Zheng, Garriga-Alonso,
  Jesson, Makar, and Blei]{circuithyptest2024}
Claudia Shi, Nicolas Beltran-Velez, Achille Nazaret, Carolina Zheng, Adri\`a
  Garriga-Alonso, Andrew Jesson, Maggie Makar, and David~M. Blei.
\newblock Hypothesis testing the circuit hypothesis in llms.
\newblock \emph{arXiv preprint arXiv:2410.13032}, 2024.

\bibitem[Wang et~al.(2026)Wang, Li, Chen, Chu, Fan, and Hu]{knowattr2026}
Bo~Wang, Junzhuo Li, Hong Chen, Yuanlin Chu, Yuxuan Fan, and Xuming Hu.
\newblock Deconstructing pre-training: Knowledge attribution analysis in moe
  and dense models.
\newblock \emph{arXiv preprint arXiv:2601.08383}, 2026.

\end{thebibliography}
\end{document}